\documentclass{article}

\usepackage[dblblindworkshop, final]{neurips_2026}
\workshoptitle{TAE (Trust-AI-Eval): Can We Trust AI Evaluation?}

\usepackage[utf8]{inputenc}
\usepackage[T1]{fontenc}
\usepackage{hyperref}
\usepackage{url}
\usepackage{booktabs}
\usepackage{array}
\usepackage{amsfonts}
\usepackage{amsmath}
\usepackage{nicefrac}
\usepackage{microtype}
\usepackage{xcolor}
\usepackage{graphicx}
\usepackage{tikz}
\usetikzlibrary{arrows.meta, positioning, calc, decorations.pathreplacing}

\newcommand{\Inc}{\textsc{include}}
\newcommand{\Exc}{\textsc{exclude}}
\newcommand{\Indet}{\textsc{indeterminate}}
\newcommand{\GemmaFull}{Gemma 4 31B}
\newcommand{\QwenFull}{Qwen 3.6 27B}
\newlength{\exColA}
\newlength{\exColB}
\newlength{\exColC}
\newlength{\exColD}

\newif\ifsonnet
\sonnettrue
\newif\ifsonnetbreadth
\sonnetbreadthtrue
\newif\ifcrosssite
\crosssitetrue
\newif\ifrepeat
\repeattrue
\newif\ifrerun
\reruntrue
\newcommand{\taeNUnitsSample}{4,908}
\newcommand{\taeNHate}{600}
\newcommand{\taeNAnalyzableProtest}{3,249}
\newcommand{\taeNCorrupt}{188}
\newcommand{\taeNInstr}{39}
\newcommand{\taeNDefs}{25}
\newcommand{\taeNDefsProtestWord}{five}
\newcommand{\taeNDefsOtherWord}{four}
\newcommand{\taeNCrossPairs}{10}
\newcommand{\taeNCrossPairsOther}{6}
\newcommand{\taeNBoot}{2,000}
\newcommand{\taeNPilot}{300}
\newcommand{\taeAllConstructWord}{six}
\newcommand{\taeArmCountWord}{three}
\newcommand{\taeArmPhrase}{three models (Gemma 4 31B, Qwen 3.6 27B, and Claude Sonnet 5)}
\newcommand{\taeGTwoDen}{50}
\newcommand{\taeGTwoGemma}{48}
\newcommand{\taeGTwoQwen}{49}
\newcommand{\taeGTwoSonnet}{48}
\newcommand{\taeGTwoBreadthDen}{88}
\newcommand{\taeGTwoBreadthGemma}{83}
\newcommand{\taeGTwoBreadthQwen}{86}
\newcommand{\taeAllTransMin}{2.9}
\newcommand{\taeAllTransMax}{61.0}
\newcommand{\taeAllFloorMin}{1.4}
\newcommand{\taeAllFloorMax}{22.2}
\newcommand{\taeAllRatioMin}{1.8}
\newcommand{\taeAllRatioMax}{7.0}
\newcommand{\taeFloorAcledGemma}{6.2}
\newcommand{\taeFloorMmGemma}{5.1}
\newcommand{\taeFloorAcledQwen}{7.5}
\newcommand{\taeFloorMmQwen}{9.9}
\newcommand{\taeFloorAcledSonnet}{16.5}
\newcommand{\taeFloorMmSonnet}{7.6}
\newcommand{\taeFloorByDefRatioMin}{2.9}
\newcommand{\taeFloorByDefRatioMax}{7.8}
\newcommand{\taePairAssnCloseLo}{19.5}
\newcommand{\taePairAssnCloseHi}{29.2}
\newcommand{\taePairAssnFarLo}{48.1}
\newcommand{\taePairAssnFarHi}{64.2}
\newcommand{\taeTransFakeSonnet}{61.0}
\newcommand{\taeFloorFakeSonnet}{22.2}
\newcommand{\taeSonnetFakeInstrWord}{six}
\newcommand{\taeSonnetFakeDegenWord}{five}
\newcommand{\taeSonnetFakeDegenShareMin}{98.8}
\newcommand{\taeSonnetFakeSplitBatchWord}{four}
\newcommand{\taeFakeBatchWord}{twelve}
\newcommand{\taeFakeBatchTransLo}{56.7}
\newcommand{\taeFakeBatchTransHi}{65.3}
\newcommand{\taeFakeBatchFloorLo}{5.6}
\newcommand{\taeFakeBatchFloorHi}{38.9}
\newcommand{\taeFakeBatchGapLo}{17.8}
\newcommand{\taeFakeBatchGapHi}{59.7}
\newcommand{\taeNestLo}{90.4}
\newcommand{\taeNestHi}{95.5}
\newcommand{\taeIncAcledGemma}{29.7}
\newcommand{\taeIncCccOldGemma}{68.6}
\newcommand{\taeIncCccNewGemma}{61.3}
\newcommand{\taeIncCntsGemma}{9.1}
\newcommand{\taeIncMmGemma}{11.4}
\newcommand{\taeIndetMinGemma}{18.6}
\newcommand{\taeIndetMaxGemma}{63.2}
\newcommand{\taeIncAcledQwen}{22.4}
\newcommand{\taeIncCccOldQwen}{66.1}
\newcommand{\taeIncCccNewQwen}{57.6}
\newcommand{\taeIncCntsQwen}{5.5}
\newcommand{\taeIncMmQwen}{9.0}
\newcommand{\taeIndetMinQwen}{10.1}
\newcommand{\taeIndetMaxQwen}{56.4}
\newcommand{\taeIncAcledSonnet}{54.4}
\newcommand{\taeIncCccOldSonnet}{72.0}
\newcommand{\taeIncCccNewSonnet}{70.3}
\newcommand{\taeIncCntsSonnet}{8.8}
\newcommand{\taeIncMmSonnet}{12.3}
\newcommand{\taeIndetMinSonnet}{8.6}
\newcommand{\taeIndetMaxSonnet}{41.8}
\newcommand{\taeIncSpanLo}{5.5}
\newcommand{\taeIncSpanHi}{72.0}
\newcommand{\taeIndetSpanLo}{8.6}
\newcommand{\taeIndetSpanHi}{63.2}
\newcommand{\taeCccAgrGemma}{90.3}
\newcommand{\taeCccDeltaGemma}{-7.3}
\newcommand{\taeCccDeltaLoGemma}{-8.5}
\newcommand{\taeCccDeltaHiGemma}{-6.1}
\newcommand{\taeCccFlipStrictGemma}{264}
\newcommand{\taeCccFlipLooseGemma}{10}
\newcommand{\taeCccAgrQwen}{87.0}
\newcommand{\taeCccDeltaQwen}{-8.5}
\newcommand{\taeCccDeltaLoQwen}{-9.7}
\newcommand{\taeCccDeltaHiQwen}{-7.3}
\newcommand{\taeCccFlipStrictQwen}{284}
\newcommand{\taeCccFlipLooseQwen}{8}
\newcommand{\taeCccAgrSonnet}{79.7}
\newcommand{\taeCccDeltaSonnet}{-1.7}
\newcommand{\taeCccDeltaLoSonnet}{-3.4}
\newcommand{\taeCccDeltaHiSonnet}{-0.1}
\newcommand{\taeCccFlipStrictSonnet}{309}
\newcommand{\taeCccFlipLooseSonnet}{14}
\newcommand{\taeCccDeltaSmall}{-1.7}
\newcommand{\taeCccDeltaLarge}{-8.5}
\newcommand{\taeShareIndTerrGemmaLo}{17.1}
\newcommand{\taeShareIndTerrGemmaHi}{71.2}
\newcommand{\taeShareIncSpanAssn}{45.2}
\newcommand{\taeShareIncSpanTerr}{23.5}
\newcommand{\taeShareIncSpanClim}{18.8}
\newcommand{\taeShareIncLowHi}{3.3}
\newcommand{\taeRepeatFloorSonnet}{8.2}
\newcommand{\taeRepeatAgrSonnet}{91.8}
\newcommand{\taeRepeatNSonnet}{182}
\newcommand{\taeRepeatChunkWordSonnet}{four}
\newcommand{\taeRepeatFloorProtestSonnet}{5.0}
\newcommand{\taeFloorSonnet}{12.0}
\newcommand{\taeTransSonnet}{47.1}
\newcommand{\taeRepeatFloorAssnSonnet}{12.2}
\newcommand{\taeFloorAssnSonnet}{12.1}
\newcommand{\taeTransAssnSonnet}{48.6}
\newcommand{\taeRepeatBreadthOkAgr}{94.0}
\newcommand{\taeRepeatBreadthOkN}{149}
\newcommand{\taeRepeatFakeN}{50}
\newcommand{\taeXSiteAgrGemma}{99.25}
\newcommand{\taeXSiteAgrQwen}{98.88}
\newcommand{\taeEffortMixedPct}{15.5}
\newcommand{\taeEffortCleanPct}{18.3}
\newcommand{\taeRerunN}{20,754}
\newcommand{\taeRerunDisagreePct}{1.05}
\newcommand{\taeRerunMaxShift}{0.5}
\newcommand{\taeGTwoBreadthQwenFirst}{53}
\newcommand{\taeMissingN}{21}
\newcommand{\taeMissingPrompts}{8}
\newcommand{\taeVerifierChecks}{250}

\title{Measuring Definition Sensitivity in LLM Annotation of Social Science Concepts}

\author{Bogdan Mamaev\thanks{Equal contribution.}\\ Deakin University\\ \texttt{bogdan.mamaev@deakin.edu.au} \And Qizhou Wang\footnotemark[1]\\ University of Melbourne\\ \texttt{mike.wang@unimelb.edu.au} \AND Matteo Vergani\\ Deakin University\\ \texttt{matteo.vergani@deakin.edu.au} \And Christopher Leckie\\ University of Melbourne\\ \texttt{caleckie@unimelb.edu.au}}

\begin{document}

\maketitle

\begin{abstract}

In social science research, text annotation is a fundamental task that is traditionally a costly, hard-to-scale manual process. Researchers have used large language models (LLMs) to perform annotation by writing a definition of a concept into a prompt – a rule that determines whether a label should be attributed to a text depending on its content. However, concepts such as hate speech, fake news, and terrorism typically have several theoretically defensible competing definitions. It is also unknown whether the model actually applied the definition to provide the label or the label was based on the model's pretrained notion of the concept. To test this, we have the model annotate the same texts twice, each time with a different definition. We call the disagreement between the two sets of annotations definition sensitivity. As a control, we paraphrase one definition but leave its requirements unchanged – the disagreement between the annotations with the two wordings becomes the paraphrase floor. Across six concepts and three LLMs, a swap of competing definitions always changes more annotations, on average, than a paraphrase. This shows that the annotations depend on the prompted definition, so any set of annotations should be interpreted together with the definition that produced it. The LLM annotator's definition sensitivity and its paraphrase floor should both be reported in such work.

\end{abstract}
\section{Introduction}
\label{sec:intro}

Datasets for complex social phenomena, such as terrorism, protest, climate contrarianism, and fake news, are the foundation of research in many fields. In political science \citep{benoit2016crowdsourced, gleditsch2002armed}, computational social science \citep{barbera2021automated, hu2022conflibert}, sociology \citep{earl2004use}, and economics \citep{allcott2017social, enders2011terrorism}, among others, this data provides the gold labels created using manual annotation. An annotator, often a domain expert or a person employed to perform the task, reads each text and applies a written set of rules (a definition) to determine whether the text records an event or a phenomenon. However, such definitions and rules are often nuanced and contested, i.e.\ disagreed upon by researchers \citep{adcock2001measurement, sambanis2004what}. This makes manual annotation more tedious and introduces subjectivity into the datasets that other researchers use.

As large language models (LLMs) increasingly supplement or replace manual text annotation, they change how such work is conducted and evaluated. Studies now raise questions about which nuances the models still miss and how they handle tasks that require different domain expertise \citep{ollion2024dangers, ziems2024transform}. Whether and how human annotators apply their annotation rules (codebooks) is a traditional question, but little is known about whether and how LLMs apply them \citep{halterman2025codebook}. Current practice is to provide a codebook definition and its attributes as a prompt, then report the model's labels as that definition's output \citep{kristensenmclachlan2023chatbots, pangakis2023automated, pangakis2025keeping}. For example, \citet{gilardi2023chatgpt} provide the model with the same codebook that their crowd workers received and report the model's labels and compare its performance with the workers'. \citet{tornberg2024best} recommends that researchers draft such a prompt codebook as a step in the annotation workflow.

Nevertheless, the assumption that the model will apply the definition is rarely tested, and it is unclear how much the definition changes the labels. Similarly to human annotators, the model already has its own notion of the concept from prior training. The literature suggests that human annotators label most clear-cut cases as the codebook prescribes, but research often focuses on grey-zone cases, where the annotators disagree because of complexity, unclear definitions, and bias. Examples include whether a labour strike that never extends beyond a single institution counts as a protest event \citep{clark2016mass, ccc2024guidelines}. Another example is whether an assassination counts as a terrorist act when the killer targets an individual \citep{enders2011terrorism, weinberg2004challenges}. Such disagreements arise from both codebook design and the human annotators who apply the codebook. Do models reproduce these disagreements, and to what degree do they annotate according to the codebook that they receive?

In this study, we measure whether a concept definition changes the model's annotations while all other parts of the prompt remain unchanged. Our design covers \taeAllConstructWord{} concepts with competing definitions from widely used datasets and studies -- protest, assassination, terrorism, hate speech, climate contrarianism, and fake news. We pair each definition with the same query. Each of the \taeArmPhrase{} annotates every text in a concept's corpus as \Inc, \Exc, or \Indet.

A comparison of the annotations made with two competing definitions on the same texts determines \emph{definition sensitivity}, the share of texts on which they disagree. However, wording can also change annotations, so we paraphrase one definition of each concept and a second protest definition, i.e.\ simply change their wording but not their rules (\taeNInstr{} prompts in total). For two wordings of one definition, the share of texts on which their annotations disagree is the \emph{paraphrase floor}. We use the floor as a baseline and compare definition sensitivity with it, as a gap in percentage points and as their ratio. That comparison shows whether the rules specified by the definitions, not only wording, changed the annotations.

Across all \taeAllConstructWord{} concepts and \taeArmCountWord{} models, definition sensitivity was \taeAllTransMin--\taeAllTransMax\% against paraphrase floors of \taeAllFloorMin--\taeAllFloorMax\%, and \taeAllRatioMin--\taeAllRatioMax{} times its floor. A definition revised by its own dataset provider also changed the annotations. A swap of the 2021 Crowd Counting Consortium definition for its 2024 successor shifted the share of protest texts labelled \Inc{} by $\taeCccDeltaSmall$ to $\taeCccDeltaLarge$ percentage points across the \taeArmCountWord{} models.

Our contribution is to measure how much the unreported choice of definition changes LLM annotations, beyond the effect of its wording. We argue that such measurements can show this effect in any annotation task where an LLM applies a definition to produce labels.
\section{Related Work}
\label{sec:related}

\noindent\textbf{Measurement in political science.} Notably, studies in political science have observed that one annotated corpus can support several incompatible but theoretically defended measures of one concept \citep{dinas2010measuring, franzmann2006locating}. Most attention is paid to how annotated information is then converted into scores \citep{lowe2011scaling}, while the step where sentences are assigned labels is rarely contested \citep{gemenis2013what, mikhaylov2012coder}. For event data, such as protests or violent clashes, published definitions tend to disagree on what counts as an instance of the event itself due to different theoretical interpretations, analytical approaches, and operationalisations. For example, the Crowd Counting Consortium's 2024 protest definition excludes labour strikes, while the Conflict and Mediation Event Observations (CAMEO) ontology explicitly includes them \citep{halterman2025codebook}. The choice to include or exclude them changes the data and possibly any study that uses it. Event datasets built from sources that overlap disagree substantially \citep{eck2012trust, oberg2025coding}, and aggregation adds another layer of variation \citep{vanbaalen2026aggregation}.

\noindent\textbf{Competing definitions of one concept.} \citet{sambanis2004what} used twelve competing definitions of civil war in their model of civil war onset to test whether conclusions about civil war depend on its definition. Each definition produced a different list of onsets, so the definition impacted the predictive power of terrain, oil, and ethnic heterogeneity in the model. Sambanis also observed that some predictors, such as income and population, remained strong across the definitions. In a different study, \citet{gibler2016analysis} rechecked which conflicts between states the Militarized Interstate Dispute data should include and how it describes them, and revised about two-thirds of the data. Satisfaction of both states with the status quo then no longer predicted three measures of escalation. However, the factors that predict whether two states start a dispute, such as a shared border, remained the same.

\citet{breznau2022observing} ask 73 teams to test with the same survey data whether immigration reduces public support for social policies. Their estimates of this effect range from large negative to large positive, and the teams' identifiable analysis decisions explain less than 5\% of these differences. \citet{silberzahn2018many} ask 29 teams whether soccer referees issue more red cards to players with darker skin, and the teams' odds ratios range from 0.89 to 2.93. \citet{menkveld2024nonstandard} note that such a spread of results is zero when all teams agree on the most reasonable analysis. Breznau's teams pick which of six survey questions measures support for social policies, and Silberzahn's pick covariates and statistical methods. Thus, these spreads mix the choice of definition with other analysis choices.

\noindent\textbf{LLM annotation.} Computational social science tends to focus on the annotation pipeline. \citet{he2026codebooks} treat codebook-guided annotation as a measurement task where a richer codebook in the prompt raises macro-F1 from 0.457 with label names only to 0.633. However, when they shuffle which label name is attached to which definition, every method evaluated by them scores below 0.20 weighted F1. \citet{mclaren2026magic} treat model choice, prompt style, and learning approach as pipeline decisions. They find that none of them has a stable effect on its own, and what works depends on the task at hand and the model. \citet{baumann2025hacking} run thirteen million annotations across configurations and conclude that each hypothesis test should use more than one. \citet{reiss2023testing} performs one annotation task with ten different instructions and reports a mean Krippendorff's alpha of 0.43 between instructions.

\noindent\textbf{Prompt and format sensitivity.} The machine learning literature tends to focus on the prompt and how it is worded and formatted. For example, \citet{sclar2024quantifying} report differences of up to 56 accuracy points across 320 prompt formats. \citet{mizrahi2024state} reword the instructions of a task and find that model rankings correlate negatively across some paraphrases in 15 of 25 tasks. \citet{alzahrani2024benchmarks} introduce small changes to the prompting format, such as a new order of the answer choices, which changes a model's leaderboard position by up to eight places. \citet{biderman2024lessons} report that three independently written implementations of the same benchmark, MMLU, produce different scores from the same models and impact their ranking order.

\noindent\textbf{Label error and model rankings.} The literature presents evidence that changes in the gold labels of benchmark test sets, such as the image classes in ImageNet, have little effect on model rankings. \citet{northcutt2021pervasive} correct label errors across benchmark test sets and find that rankings on the full sets remain unchanged, and models reorder only on the subset of items where corrected and original labels disagree. \citet{recht2019generalize} rebuild test sets from scratch with the original collection procedure and find that every model loses accuracy, but their ranking order stays similar. \citet{beyer2020imagenet} relabel the same test set and find that the differences between models shrink rather than reverse.

When machine learning studies replace one set of gold labels with another, they tend to declare the new set as an improvement over the initial one. \citet{beyer2020imagenet} present their relabelling as the more accurate measurement, so their two label sets are treated as a mistake and its correction. Construct-validity critiques of benchmarking \citep{raji2021everything} also ask researchers to rank label sets by how well they capture the concept and use the better one \citep{bowman2021benchmarking}.

\noindent\textbf{Prediction-powered inference.} Prediction-powered frameworks tend to address label error rather than the choice of definition. They use a small expert-labelled sample to debias estimates from machine labels, and their confidence intervals stay valid however accurate those labels are \citep{angelopoulos2023prediction, egami2023dsl, egami2024general, zrnic2024cross}. These frameworks assume that documents are drawn for gold labels with a sampling probability known to researchers and ``larger than zero for every document'' \citep[p.~19]{egami2024general}. They also assume that an expert who reads a sampled text can assign its true label. However, this assumption can fail when a text does not contain information the definition requires. For example, if a definition requires a crowd size, the label cannot be reliably assigned if the size is not mentioned in the text. One way to handle such cases is design-based supervised learning, which takes the vote of a panel of experts as the answer and adds their disagreement to the estimate's uncertainty \citep{egami2024general}.

We address four questions that these studies raise:
\begin{enumerate}
\item How does a score computed from annotated text change when a defended competing definition provides the rules? \citet{sambanis2004what} ran such a design by hand, before LLMs and without a baseline to compare the spread with.
\item How much does a paraphrase of a definition change the labels? \citet{biderman2024lessons} report a comparable difference between three implementations of one benchmark.
\item What should the correct label be when a text does not contain what the definition requires? The prediction-powered frameworks tend to resolve this question by a rule, and \citet{martinon2026glide} note, as an open problem, that expert disagreement can change what is estimated.
\item Does the model apply the definition at all? \citet{lemens2025positioning} ask models to place parties on a policy dimension without a definition, and the best models' scores correlate above 0.90 with expert ratings. \citet{halterman2025codebook} find that, once the other parts of the codebook are removed, zero-shot F1 rises when the class definition is dropped too.
\end{enumerate}
\section{Evaluation Setup}
\label{sec:design}
\begin{figure}[t]
\centering
\resizebox{\linewidth}{!}{%
\begin{tikzpicture}[
  font=\footnotesize,
  tbox/.style={draw=black!80, line width=0.5pt, rounded corners=2.5pt, fill=white,
               align=center, minimum height=4.2em, inner sep=3.5pt},
  sub/.style={font=\scriptsize, text=black!55},
  stage/.style={font=\footnotesize\itshape, text=black!55, anchor=south},
  arr/.style={-{Stealth[length=5pt, width=4.5pt]}, draw=black!70, line width=0.6pt,
              rounded corners=6pt},
]
\draw[draw=black!40, dashed, rounded corners=5pt, fill=black!3]
  (2.15,-1.25) rectangle (9.78,1.28);
\node[stage, anchor=center] at (5.97, 1.06) {held constant};
\node[sub] at (5.97, -1.02) {same texts \enspace$\cdot$\enspace same model};
\node[tbox, text width=8.6em] (defs) at (0, 1.15)
  {competing definitions\\[1pt]{\scriptsize\color{black!55} $d_1,\dots,d_k$,\\ quoted from\\ codebooks and studies}};
\node[tbox, text width=8.6em] (paras) at (0, -1.15)
  {paraphrases of $d_1$\\[1pt]{\scriptsize\color{black!55} wording changed,\\ rule preserved}};
\node[tbox, text width=8.8em] (tmpl) at (4.05, 0.05)
  {one prompt per wording\\[1pt]{\scriptsize\color{black!55} only the quoted text differs}};
\node[tbox, text width=8.8em] (cls) at (7.93, 0.05)
  {annotations\\[1pt]{\scriptsize\color{black!55} \Inc{}, \Exc{},\\ \Indet}};
\node[tbox, anchor=west, text width=8.4em] (sens) at (10.42, 1.15)
  {definition sensitivity\\[1pt]{\scriptsize\color{black!55} cross-definition pairs}};
\node[tbox, anchor=west, text width=8.4em] (floor) at (10.42, -1.15)
  {paraphrase floor\\[1pt]{\scriptsize\color{black!55} wording pairs}};
\draw[arr] (defs.east) -- ++(0.38,0) |- (tmpl.west);
\draw[arr] (paras.east) -- ++(0.38,0) |- (tmpl.west);
\draw[arr] (tmpl.east) -- (cls.west);
\draw[arr] (cls.east) -- ++(0.42,0) |- (sens.west);
\draw[arr] (cls.east) -- ++(0.42,0) |- (floor.west);
\draw[decorate, decoration={brace, amplitude=4pt}, draw=black!75]
  ($(sens.north east)+(1.2mm,0)$) -- ($(floor.south east)+(1.2mm,0)$)
  node[midway, right=5.5pt, align=center] {gap,\\ ratio};
\node[stage] at (defs.north |- 0,2.05) {varied};
\node[stage] at (sens.north |- 0,2.05) {quantities};
\end{tikzpicture}}
\caption{Study design. Competing definitions of one concept, and paraphrases of one definition with the rule preserved, are each paired with the same query in a separate prompt. The same model annotates the same texts with every prompt and returns \Inc{}, \Exc{}, or \Indet{}. The mean pairwise disagreement over cross-definition pairs is definition sensitivity, and the mean over wording pairs of one definition is the paraphrase floor. We report their difference as a gap and their quotient as a ratio.}
\label{fig:design}
\end{figure}
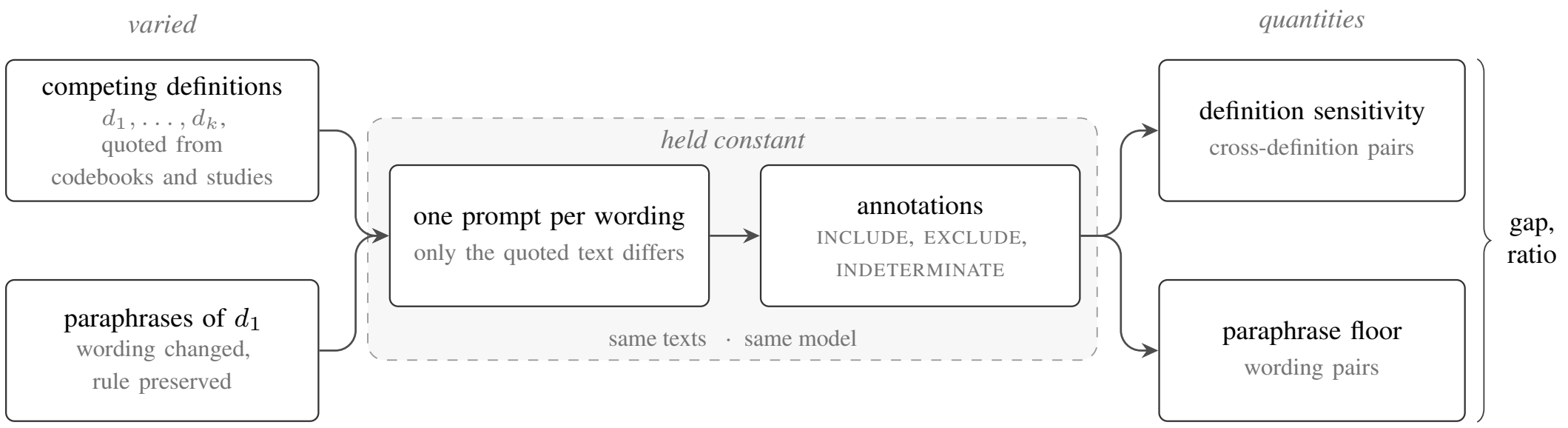

We investigate whether LLM annotators are guided by the definition that researchers write into the prompt. To test this, we measure how the annotation outcome changes with respect to competing definitions. Here, we replace only the definition and keep the other parts of the prompt unchanged. We further measure the paraphrase floor to isolate the effect of changes in wording. We conduct pairwise comparisons of the annotations from two prompts on the same texts. Figure~\ref{fig:design} summarises the pipeline.

\subsection{Dataset Information}
\label{sec:design-sample}

Our sample consists of \taeNUnitsSample{} news-derived texts for protest and assassination, plus \taeNHate{} texts for each of hate speech, climate contrarianism, and fake news. We drew protest texts from the event notes of the Armed Conflict Location and Event Dataset (ACLED) for the United States, Canada, and Mexico. We also used Crowd Counting Consortium (CCC) event records for the United States. The other protest texts are election-period news articles fetched from URLs indexed by the Global Database of Events, Language, and Tone (GDELT). The assassination texts are news articles fetched from URLs indexed by GDELT, and we reuse them for terrorism so that only the definition changes.

The protest subsample combines ordinary cases with grey-zone ones, where the disagreement often happens over, for example, whether a labour action, a target, an electoral rally, violence, or a small crowd counts. The assassination subsample is structured similarly due to possible disagreements over failed attempts, the role of the victim, whether the killing must also be terrorism, and killings by the state or in custody. The texts for hate speech \citep{borkan2019nuanced}, climate contrarianism \citep{diggelmann2020climate}, and fake news \citep{wang2017liar} are drawn randomly.

\subsection{Prompts}
\label{sec:design-instruments}

Each prompt pairs a definition with a query. Each definition consists of several rules that were drawn from the codebooks and studies for protest \citep{raleigh2010introducing, acled2026codebook, ccc2024guidelines, banks2024cnts, clark2016mass}, assassination \citep{gtd2021codebook, banks2024cnts, icrc2026targeted}, terrorism \citep{gtd2021codebook, schmid2011revised, enders2011terrorism, weinberg2004challenges}, hate speech \citep{davidson2017automated, waseem2016hateful, founta2018large, degibert2018hate}, climate contrarianism \citep{coan2021computer, cook2013quantifying, capstick2014what, rahmstorf2004climate}, and fake news \citep{lazer2018science, allcott2017social, gelfert2018fake, zhou2020survey}. Everything outside the quoted rules remained the same across the prompts of one concept.

In total, we created \taeNInstr{} prompts from \taeNDefs{} definitions across the \taeAllConstructWord{} concepts. Of those definitions, \taeNDefsProtestWord{} are for protest and \taeNDefsOtherWord{} for each of the others. Seven definitions were paraphrased twice to establish the paraphrase floor. Appendix~\ref{app:repro} reports how we froze the prompts and released their hashes. 

\subsection{LLM Annotation}
\label{sec:design-arms}

For annotation, we locally deployed \GemmaFull{} (Gemma) and \QwenFull{} (Qwen) at temperature 0 with a seed set in advance. Claude Sonnet 5 (Sonnet) was used without temperature control and was provided with the texts in batches of at most 50. Therefore, we calculated the \emph{repeat floor}, which is the share of Sonnet's annotations that change when the same prompt is run again on the same texts. Each label can assume one of three values: \Inc, \Exc, or \Indet. \Indet{} is used when a text does not contain what the definition requires (e.g., when the rule depends on a crowd size and the text does not report it). Failed requests were treated as missing data and were not converted into a label.

\subsection{Evaluation Protocol}
\label{sec:design-estimands}
We denote the annotation that a model $M$ assigns to a text $u$ when prompted with
definition $d$ as $y^{M}_{(u,d)}$. Here $D_{c}$ is the predefined finite set of
competing definitions of a concept $c$ that we analyse. For each $d \in D_{c}$,
$R_{c}(d)$ is the set of its wordings, the original and its paraphrases, which all
preserve its rule. Let $U_{c}$
be the corpus for a concept $c$ and $U_c^{(d,d')} \subseteq U_c$ the texts
annotated with both $d$ and $d'$. The two annotations disagree when
$y^M_{(u,d)} \neq y^M_{(u,d')}$, and the disagreement rate for that
pair is
\begin{equation}
\label{eq:delta-pair}
\delta^M_{(d,d')} \;=\; \frac{1}{|U_c^{(d,d')}|}
\sum_{u \in U_c^{(d,d')}} \mathbf{1}\!\left[ y^M_{(u,d)} \neq y^M_{(u,d')} \right].
\end{equation}

Let $P^{\mathrm{def}}_c$ be the pairs of distinct competing definitions of a
concept $c$. Let $P^{\mathrm{par}}_c$ be the pairs of wordings drawn from
$R_c(d)$, combined across all the paraphrased definitions. Both definition sensitivity and the
paraphrase floor are the mean of $\delta^M_{(d,d')}$ over such a set $P$ of
pairs,
\begin{equation}
\label{eq:delta-mean}
\delta^{M}(P) \;=\; \frac{1}{|P|}\sum_{(d,d')\in P} \delta^{M}_{(d,d')}.
\end{equation}
Definition sensitivity is $\delta^{M}(P^{\mathrm{def}}_{c})$ over the competing definitions of one concept, \taeNCrossPairs{} pairs for protest and \taeNCrossPairsOther{} for each of the other concepts. The paraphrase floor is $\delta^{M}(P^{\mathrm{par}}_{c})$ over pairs of wordings of the same definition. We report the difference between definition sensitivity and the paraphrase floor as a gap in percentage points, and their quotient as a ratio.

For inference, we use a block bootstrap, in which the texts of a block are resampled together. For protest, assassination, and terrorism, a block contains the texts with the same calendar week, place, and trigger word, a keyword stem such as \textit{picket}. The place is recorded only for the ACLED and CCC texts. For fake news, a block contains the claims of one speaker, the person or organisation that made the claims, and each hate speech or climate contrarianism text is its own block. We perform \taeNBoot{} resamples and recompute definition sensitivity, the paraphrase floor, and their gap on each resample. We also recompute the change in the share of protest texts labelled \Inc{} between the CCC 2021 and CCC 2024 definitions. We compute 95\% percentile confidence intervals for these quantities, while the ratios and label shares are provided as point estimates.

As a comprehension check, we showed each model a definition without text for labelling and asked what that definition's rule required. We scored the answer to each question about the rule against a prepared set of acceptable answers. Gemma and Qwen scored \taeGTwoGemma{} and \taeGTwoQwen{} of \taeGTwoDen{} questions on the protest and assassination prompts. On the other four concepts, they scored \taeGTwoBreadthGemma{} and \taeGTwoBreadthQwen{} of \taeGTwoBreadthDen{}, respectively.\ifsonnet{} Sonnet scored \taeGTwoSonnet{} of \taeGTwoDen{} on the protest and assassination questions.\fi{}

Before undertaking the main part of the experiment, we ran the design on \taeNPilot{} protest texts to test whether a swap of the definition changed how many texts were labelled \Inc{}. We then pre-registered the sample, definitions, quantities to be measured, and analysis. Section~\ref{sec:limitations} and Appendix~\ref{app:deviations} report the changes introduced after pre-registration.

\section{Results}
\label{sec:results}

\subsection{Definition Sensitivity Across Concepts}
\label{sec:res-all}

Definition sensitivity exceeds the paraphrase floor in every concept and every model. Across the six concepts, it ranges from \taeAllTransMin\% to \taeAllTransMax\% against floors of \taeAllFloorMin--\taeAllFloorMax\%. Table~\ref{tab:res-transmission} reports every sensitivity, floor, and gap. The ratio is highest for protest in Gemma and Qwen. Hate speech has the lowest definition sensitivity, but the confidence interval of its gap still excludes zero in every model.

For protest, two definitions were paraphrased, so we computed the paraphrase floor of each definition separately. The floors for ACLED and Mass Mobilization Protest Data (MMPD) are \taeFloorAcledGemma\% and \taeFloorMmGemma\% for Gemma and \taeFloorAcledQwen\% and \taeFloorMmQwen\% for Qwen. For Sonnet, they are \taeFloorAcledSonnet\% and \taeFloorMmSonnet\%. In every model, definition sensitivity is \taeFloorByDefRatioMin{} to \taeFloorByDefRatioMax{} times either floor. Definition sensitivity is a mean over pairs of definitions, and the disagreement rates of the pairs differ within each concept (Table~\ref{tab:pairs}). For assassination, the closest pair, the Global Terrorism Database (GTD) and GTD with successful attacks only, disagrees on \taePairAssnCloseLo--\taePairAssnCloseHi\% of the texts across models. The farthest pair, GTD and the definition of targeted killing by the \citeauthor{icrc2026targeted}, disagrees on \taePairAssnFarLo--\taePairAssnFarHi\%.

\begin{table}[t]
\centering
\small
\caption{Definition sensitivity against the paraphrase floor, by concept and model. Each sensitivity and floor cell is the share of texts on which two annotations disagree. The rate is averaged over pairs of competing definitions for sensitivity, and over pairs of wordings of one definition for the floor. Gap is definition sensitivity minus the floor, in percentage points. Ratio is definition sensitivity divided by the floor.}
\label{tab:res-transmission}\label{tab:res-breadth}
\begin{tabular}{@{}llcccc@{}}
\toprule
& & Definition & Paraphrase & & \\
Concept & Model & sensitivity & floor & Gap & Ratio \\
\midrule
Protest & Gemma 4 31B & 39.5 & 5.6 & 33.8 & 7.0 \\
 & Qwen 3.6 27B & 44.1 & 8.7 & 35.4 & 5.1 \\
 & Claude Sonnet 5 & 47.1 & 12.0 & 35.1 & 3.9 \\
\addlinespace
Assassination & Gemma 4 31B & 53.7 & 11.7 & 42.0 & 4.6 \\
 & Qwen 3.6 27B & 40.3 & 15.2 & 25.1 & 2.7 \\
 & Claude Sonnet 5 & 48.6 & 12.1 & 36.4 & 4.0 \\
\addlinespace
Terrorism & Gemma 4 31B & 41.1 & 13.8 & 27.3 & 3.0 \\
 & Qwen 3.6 27B & 22.1 & 10.9 & 11.2 & 2.0 \\
 & Claude Sonnet 5 & 38.8 & 13.9 & 24.9 & 2.8 \\
\addlinespace
Hate speech & Gemma 4 31B & 3.4 & 1.6 & 1.8 & 2.2 \\
 & Qwen 3.6 27B & 5.0 & 2.2 & 2.8 & 2.2 \\
 & Claude Sonnet 5 & 2.9 & 1.4 & 1.5 & 2.0 \\
\addlinespace
Climate contrarianism & Gemma 4 31B & 19.9 & 7.6 & 12.3 & 2.6 \\
 & Qwen 3.6 27B & 30.4 & 11.8 & 18.6 & 2.6 \\
 & Claude Sonnet 5 & 21.0 & 11.8 & 9.2 & 1.8 \\
\addlinespace
Fake news & Gemma 4 31B & 17.2 & 9.7 & 7.6 & 1.8 \\
 & Qwen 3.6 27B & 14.3 & 4.9 & 9.4 & 2.9 \\
 & Claude Sonnet 5 & 61.0 & 22.2 & 38.8 & 2.7 \\
\bottomrule
\end{tabular}

\end{table}

\begin{table}[t]
\centering
\small
\caption{Disagreement rate of the closest and the farthest pair of definitions in each concept, in percent of the texts. Closest and farthest are by the mean rate across the three models.}
\label{tab:pairs}
\begin{tabular}{@{}ll>{\raggedright\arraybackslash}p{5.2cm}ccc@{}}
\toprule
Concept & Pair & Definitions & Gemma & Qwen & Sonnet \\
\midrule
Protest & Closest & CCC 2021, CCC 2024 & 9.7 & 13.0 & 20.3 \\
 & Farthest & CCC 2021, Cross-National Time-Series Data Archive (CNTS) & 60.8 & 71.9 & 70.3 \\
\addlinespace
Assassination & Closest & GTD, GTD with successful attacks only & 25.4 & 19.5 & 29.2 \\
 & Farthest & GTD, \citeauthor{icrc2026targeted} & 64.2 & 48.1 & 57.1 \\
\addlinespace
Terrorism & Closest & \citeauthor{enders2011terrorism}, GTD & 32.0 & 18.4 & 24.1 \\
 & Farthest & \citeauthor{schmid2011revised}, \citeauthor{weinberg2004challenges} & 58.1 & 27.0 & 52.3 \\
\addlinespace
Hate speech & Closest & \citeauthor{davidson2017automated}, \citeauthor{founta2018large} & 2.8 & 3.3 & 1.5 \\
 & Farthest & \citeauthor{degibert2018hate}, \citeauthor{waseem2016hateful} & 4.7 & 7.0 & 4.3 \\
\addlinespace
Climate contrarianism & Closest & \citeauthor{capstick2014what}, \citeauthor{coan2021computer} & 18.5 & 21.0 & 13.7 \\
 & Farthest & \citeauthor{capstick2014what}, \citeauthor{cook2013quantifying} & 20.8 & 41.5 & 28.0 \\
\addlinespace
Fake news & Closest & \citeauthor{gelfert2018fake}, \citeauthor{zhou2020survey} & 5.3 & 3.8 & 0.0 \\
 & Farthest & \citeauthor{allcott2017social}, \citeauthor{zhou2020survey} & 16.8 & 25.7 & 99.7 \\
\bottomrule
\end{tabular}

\end{table}

Sonnet's definition sensitivity for fake news, \taeTransFakeSonnet\%, is the highest in this paper, but it does not show the definitions' effect on individual texts. The texts are single-sentence claims, and they do not contain the facts about a news outlet that a definition such as that of \citet{lazer2018science} requires. Consistent with this, Sonnet assigns a single label to at least \taeSonnetFakeDegenShareMin\% of the texts with each of \taeSonnetFakeDegenWord{} of the \taeSonnetFakeInstrWord{} fake news prompts. It labels these texts \Exc{} with the definition of \citet{allcott2017social} and both paraphrases of the definition of \citet{lazer2018science}, and \Indet{} with the definitions of \citet{gelfert2018fake} and \citet{zhou2020survey}. When the single labels of two prompts are different, their annotations disagree on almost every text. Therefore, the definition sensitivity here depends on which label Sonnet assigns with each prompt.

With one prompt, Sonnet's label also varies arbitrarily between batches of at most 50 texts. With the original wording of the definition of \citet{lazer2018science}, Sonnet labels the texts of \taeSonnetFakeSplitBatchWord{} of the \taeFakeBatchWord{} batches \Indet{} and the other texts \Exc{}. With both paraphrases, Sonnet labels every text \Exc{}. So the paraphrase floor of \taeFloorFakeSonnet\% is due only to these \taeSonnetFakeSplitBatchWord{} batches, because the annotations with the three wordings agree on all other texts. When Sonnet labelled one of these batches a second time with the same prompt, all \taeRepeatFakeN{} labels changed from \Indet{} to \Exc{}. Thus, Sonnet assigns the label to a whole batch at once, and the effective sample size is \taeFakeBatchWord{} batches rather than \taeNHate{} texts.

Our bootstrap for fake news resamples speakers with all their claims, and each batch contains the claims of many speakers. When we resample the \taeFakeBatchWord{} batches instead, the interval for definition sensitivity is [\taeFakeBatchTransLo, \taeFakeBatchTransHi]. The intervals for the floor and the gap are [\taeFakeBatchFloorLo, \taeFakeBatchFloorHi] and [\taeFakeBatchGapLo, \taeFakeBatchGapHi].

\subsection{Examples of Label Changes}
\label{sec:res-examples}

Table~\ref{tab:examples} describes four texts from the sample and illustrates how one difference between two definitions changes a label. On each text, all three models returned \Inc{} with one definition and \Exc{} with the other. The rationale that each model returned with its \Exc{} label named the clause that distinguishes the two definitions. We describe the texts in our own words, because the source texts are licensed (Appendix~\ref{app:repro}). The definition in the third row, GTD with successful attacks only, adds one requirement to GTD. Therefore, every text that meets this stricter definition also meets GTD and should be labelled \Inc{} with both. Of the assassination texts that a model labelled \Inc{} with the stricter definition, \taeNestLo--\taeNestHi\% were also \Inc{} with GTD.

\begin{table}[t]
\centering
\small
\caption{Four texts on which all three models changed their label with the definition. Each row names the clause that differs and what the models' rationales cite for the \Exc{} label.}
\label{tab:examples}
\begin{tabular}{@{}>{\raggedright\arraybackslash}p{\exColA}>{\raggedright\arraybackslash}p{\exColB}>{\raggedright\arraybackslash}p{\exColC}>{\raggedright\arraybackslash}p{\exColD}@{}}
\toprule
Text & Labels & Difference & Reasons for \Exc{} \\
\midrule
An event note on about a dozen people who protested outside a high school to ask a school board member to resign. & ACLED \Inc{}, MMPD \Exc{} & ACLED counts a demonstration of three or more participants. MMPD requires 50 or more people. & All three cite the minimum of 50 people. \\
\addlinespace
An event note on dozens of nurses who picketed a hospital at the start of a strike for staffing and pay. & CCC 2021 \Inc{}, CCC 2024 \Exc{} & The 2024 edition no longer covers labour strikes or the pickets of striking workers. & All three cite the exclusion of workers' pickets. \\
\addlinespace
A news report on militants who tried to kill a government minister. The minister was stabbed but survived. & Global Terrorism Database (GTD) \Inc{}, GTD with successful attacks only \Exc{} & The second definition counts assassinations only when the target is killed. & All three cite the survival of the minister. \\
\addlinespace
A news report on gunmen who shot dead a senior army general outside his home. A newly formed militant group claimed responsibility for the attack online. & \citeauthor{weinberg2004challenges} \Inc{}, \citeauthor{schmid2011revised} \Exc{} & \citeauthor{schmid2011revised} states that the main direct victims of terrorism are in general civilians, not armed forces. \citeauthor{weinberg2004challenges} do not distinguish combatants from non-combatants. & All three cite the military role of the victim. \\
\bottomrule
\end{tabular}
\end{table}

\subsection{Event Counts}
\label{sec:res-shares}

\begin{figure}[t]
\centering
\includegraphics[width=.48\textwidth]{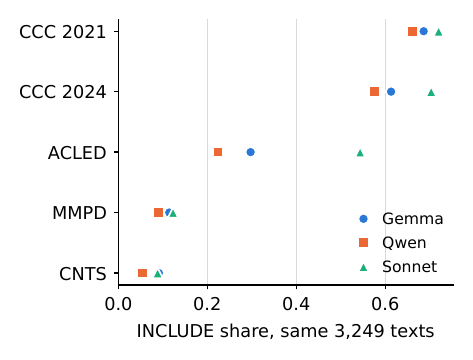}\hfill
\includegraphics[width=.48\textwidth]{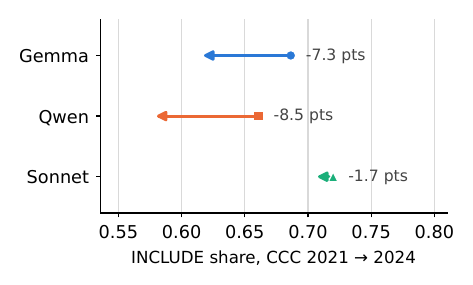}
\caption{Definitions change label shares on the same texts. In the left panel, the share of the \taeNAnalyzableProtest{} texts labelled \Inc{} is shown for each protest definition. The right panel shows the change in this share per model from CCC 2021 to CCC 2024, in percentage points.}
\label{fig:res-span-ccc}
\end{figure}

Definitions also change how many texts are labelled as events. Every model received the same \taeNAnalyzableProtest{} texts with each of the \taeNDefsProtestWord{} protest definitions, and how many it counted as \Inc{} changed with the definition (Figure~\ref{fig:res-span-ccc}, left). The smallest share any definition produced in any model was \taeIncSpanLo\% and the largest was \taeIncSpanHi\%. We quoted those \taeNDefsProtestWord{} definitions from the Cross-National Time-Series Data Archive (CNTS), MMPD, ACLED, and the CCC. The CCC has both the 2021 and the 2024 definitions (Figure~\ref{fig:res-span-ccc}, right).

In every model, the \Inc{} share rises in the same order, from CNTS through MMPD, ACLED, and CCC 2024 to CCC 2021. For CNTS, MMPD, ACLED, and CCC 2021, the shares were \taeIncCntsGemma\%, \taeIncMmGemma\%, \taeIncAcledGemma\%, and \taeIncCccOldGemma\% in Gemma. They were \taeIncCntsQwen\%, \taeIncMmQwen\%, \taeIncAcledQwen\%, and \taeIncCccOldQwen\% in Qwen\ifsonnet, and \taeIncCntsSonnet\%, \taeIncMmSonnet\%, \taeIncAcledSonnet\%, and \taeIncCccOldSonnet\% in Sonnet\fi. These shares differ more between definitions than between models. The \Inc{} share also changes with the definition for assassination, terrorism, and climate contrarianism (Table~\ref{tab:shares}). Within one model, \Inc{} shares differ between definitions by up to \taeShareIncSpanAssn{} percentage points for assassination, \taeShareIncSpanTerr{} for terrorism, and \taeShareIncSpanClim{} for climate contrarianism. For hate speech and fake news, every model labels at most \taeShareIncLowHi\% of the texts \Inc{} with any definition.

\begin{table}[t]
\centering
\small
\caption{Range of the \Inc{} and \Indet{} shares across the definitions of each concept, in percent of the texts, for the original wording of each definition.}
\label{tab:shares}
\begin{tabular}{@{}lcccccc@{}}
\toprule
& \multicolumn{3}{c}{\Inc{} share} & \multicolumn{3}{c}{\Indet{} share} \\
\cmidrule(lr){2-4}\cmidrule(l){5-7}
Concept & Gemma & Qwen & Sonnet & Gemma & Qwen & Sonnet \\
\midrule
Protest & 9.1--68.6 & 5.5--66.1 & 8.8--72.0 & 18.6--63.2 & 10.1--56.4 & 8.6--41.8 \\
Assassination & 22.8--68.0 & 13.4--41.1 & 25.6--60.5 & 10.3--22.4 & 2.3--13.1 & 7.7--19.4 \\
Terrorism & 8.7--23.2 & 11.9--28.3 & 29.0--52.5 & 17.1--71.2 & 8.2--20.9 & 10.7--47.0 \\
Hate speech & 0.3--2.8 & 0.3--2.3 & 0.7--3.3 & 0.0--1.5 & 1.8--5.2 & 0.0--0.5 \\
Climate contrarianism & 18.0--30.7 & 13.8--25.2 & 28.0--46.8 & 4.0--19.3 & 5.5--35.5 & 10.3--12.5 \\
Fake news & 0.0--2.8 & 0.0--0.8 & 0.0--0.8 & 77.2--96.5 & 74.7--99.3 & 0.3--100.0 \\
\bottomrule
\end{tabular}

\end{table}

CCC 2021 and CCC 2024 are two editions of one published codebook, both among the protest definitions above. The 2024 edition is stricter, and every model counts fewer texts as \Inc{} under it (Figure~\ref{fig:res-span-ccc}, right). The label is identical under both editions for \taeCccAgrGemma\% of the same \taeNAnalyzableProtest{} texts in Gemma, \taeCccAgrQwen\% in Qwen\ifsonnet, and \taeCccAgrSonnet\% in Sonnet\fi. The \Inc{} share falls from \taeIncCccOldGemma\% to \taeIncCccNewGemma\% in Gemma ($\taeCccDeltaGemma$ percentage points [$\taeCccDeltaLoGemma$, $\taeCccDeltaHiGemma$]) and from \taeIncCccOldQwen\% to \taeIncCccNewQwen\% in Qwen ($\taeCccDeltaQwen$ [$\taeCccDeltaLoQwen$, $\taeCccDeltaHiQwen$]).\ifsonnet{} In Sonnet, it falls from \taeIncCccOldSonnet\% to \taeIncCccNewSonnet\% ($\taeCccDeltaSonnet$ [$\taeCccDeltaLoSonnet$, $\taeCccDeltaHiSonnet$]).\fi{} The intervals of Gemma and Qwen exclude zero.\ifsonnet{} Sonnet's upper bound is so close to zero that whether its interval excludes zero depends on the random seed of the bootstrap.\fi{}

Nearly all changes between \Inc{} and \Exc{} are in the stricter direction. Gemma switches \taeCccFlipStrictGemma{} labels from \Inc{} to \Exc{} and \taeCccFlipLooseGemma{} back, Qwen \taeCccFlipStrictQwen{} and \taeCccFlipLooseQwen{}\ifsonnet, and Sonnet \taeCccFlipStrictSonnet{} and \taeCccFlipLooseSonnet{}\fi. The remaining changes are to or from \Indet{}.

\subsection{Abstention}
\label{sec:res-abstention}

Across the \taeNDefsProtestWord{} protest definitions, the shares of \Indet{} labels span \taeIndetMinGemma--\taeIndetMaxGemma\% in Gemma, \taeIndetMinQwen--\taeIndetMaxQwen\% in Qwen\ifsonnet, and \taeIndetMinSonnet--\taeIndetMaxSonnet\% in Sonnet\fi. Every model assigns \Indet{} least often with the two CCC definitions and most often with CNTS or MMPD. This difference emerges because definitions require different facts, and news reports often omit them, which results in a text that is \Indet{} by one definition and \Inc{} by another. The share of \Indet{} labels also changes with the definition in other concepts, for example from \taeShareIndTerrGemmaLo\% to \taeShareIndTerrGemmaHi\% for terrorism in Gemma (Table~\ref{tab:shares}).
\section{Conclusion}
\label{sec:implications}

In every concept that we tested, a swap of a concept definition for a competing one changed more labels, on average, than its paraphrase. The size of this change varied, so the applied definition, its prompt, and the paraphrase floor need to be reported, not only the name of the phenomenon. A change of definition may affect how results are interpreted, and this information complements the evaluative metrics for models and the procedural reports for datasets \citep{gebru2021datasheets, mitchell2019model}. Applying even slightly different definitions to the same texts is likely to produce different results.

Our results answer the four questions of Section~\ref{sec:related}. First, definition sensitivity, the mean disagreement rate between competing definitions, was \taeAllTransMin\% to \taeAllTransMax\% across concepts and models. The share of texts labelled \Inc{} also changed with the definition, for example by up to \taeShareIncSpanAssn{} percentage points within one model for assassination. Second, the paraphrase floor, the mean disagreement rate between paraphrases of one definition, was \taeAllFloorMin\% to \taeAllFloorMax\%. Third, the share of \Indet{} labels changed with the definition (Table~\ref{tab:shares}). For the \Indet{} texts, a binary label requires a rule that the definition does not state. Fourth, definition sensitivity exceeded the paraphrase floor in every concept and model, which suggests that the models applied the definition in the prompt.

Reporting definition sensitivity and the paraphrase floor can help researchers understand how their definition affects the data, and it can make reports on annotation choices more informative without the need to build new gold labels. Using definition sensitivity is also informative when trying to understand how the evaluator's own definition compares with competing definitions, and to establish where the difference emerges. Studies of prompt sensitivity tend to vary the wording of a single rule \citep[e.g.][]{mizrahi2024state}, and we recommend that evaluations also vary which rule is used.

When definition sensitivity is well above the paraphrase floor, this shows that the definition affected the labels that the model produced. Conversely, when definition sensitivity is close to the paraphrase floor, the annotations are likely to depend on the model's pretrained notion of the concept. A study that presents those annotations as the codebook's output may then report this notion. A high paraphrase floor, regardless of the definition sensitivity, shows that a paraphrase changes many annotations, which warrants further investigation and adjustments to the prompt and possibly to the definitions. In any case, it is informative to report both measures.

We have also found that revisions of codebooks may change labels, even if the changes are perceived as marginal. It would be good practice to measure how changes in a codebook affect the labels. When the Crowd Counting Consortium tightened its protest rule from CCC 2021 to CCC 2024, every model labelled fewer texts \Inc{} with the later edition. Unchanged labels would have suggested that the models did not apply the revised rule. Social scientists have run such comparisons by hand \citep[e.g.][]{gibler2016analysis}, and LLM annotation can make them less expensive.

Definitions are different in what evidence they require from a text. For example, on the protest corpus, the share of \Indet{} labels ranges from \taeIndetSpanLo{}\% to \taeIndetSpanHi{}\% across definitions and models. The highest share is that of Gemma with CNTS, and with this definition Gemma labels \taeIncCntsGemma\% of the protest texts \Inc{}. For binary gold labels, those \Indet{} texts have to be labelled \Inc{} or \Exc{}, and the rule for this choice becomes part of the gold labels. For Gemma with CNTS, this rule alone can change the \Inc{} share by up to \taeIndetSpanHi{} percentage points. We therefore suggest that studies report the share of \Indet{} labels for each definition, with the rule that assigns these texts to \Inc{} or \Exc{}.

An evaluation can also report one concept-level score averaged over competing definitions. Competing definitions can correspond to different estimands, or quantities to estimate, each of them theoretically defensible. Their disagreement on the protest corpus is also structured, with the same order of \Inc{} shares in every model (Section~\ref{sec:res-shares}). Which definition is correct depends on the estimand. However, definition sensitivity measures how much the choice of definition changes the labels, and an averaged score does not show this.
\section{Limitations}
\label{sec:limitations}

We drew the hate speech, climate contrarianism, and fake news texts at random from public datasets. By contrast, we chose the protest and assassination texts to include grey-zone cases. Therefore, definition sensitivity for protest, assassination, and terrorism may be higher because of this selection, so the values should be compared with caution between concepts. Results on \taeAllConstructWord{} concepts, with texts in English and Spanish, also do not show that definition sensitivity exceeds the paraphrase floor for other concepts or languages.

\ifsonnet Sonnet's repeat floor is specific to the command-line harness that sent its prompts (Appendix~\ref{app:deviations}).\fi\ifrepeat{} After labelling finished, we re-ran \taeRepeatChunkWordSonnet{} batches of protest and assassination texts, and Sonnet reproduced \taeRepeatAgrSonnet\% of \taeRepeatNSonnet{} annotations, a repeat floor of \taeRepeatFloorSonnet\%. On protest, the repeat floor is \taeRepeatFloorProtestSonnet\%, against Sonnet's paraphrase floor of \taeFloorSonnet\% and definition sensitivity of \taeTransSonnet\%. On assassination, it is \taeRepeatFloorAssnSonnet\%, against \taeFloorAssnSonnet\% and \taeTransAssnSonnet\%.\fi{} The open-weight models ran at temperature 0 with a seed set in advance, and we assume that their repeat floor is zero. We did not run the pre-registered check that a re-run on the same platform reproduces every label.\ifcrosssite{} A re-run on a second platform reproduced \taeXSiteAgrGemma\% of Gemma's and \taeXSiteAgrQwen\% of Qwen's labels on the other four concepts.\fi{} We also did not report some pre-registered analyses, such as a disagreement rate for each attribute on which definitions differ (Appendix~\ref{app:deviations}).

Our findings are specific to the definitions used and the model versions. The labels reported here evaluate model outputs; we cannot attest to labelling accuracy or whether any text describes a protest, assassination, hate speech, or fake news. The texts were written by third parties.
\label{lastcontentpage}
\begin{ack}
This work was partially supported by the Australian Internet Observatory (AIO), a national research infrastructure created to facilitate research on digital platforms and data for researchers in the humanities, social sciences, and other fields. The Australian Internet Observatory (\url{https://doi.org/10.25956/twvn-ca19}) is a co-investment partnership with the Australian Research Data Commons (ARDC) through the HASS and Indigenous RDC (\url{https://doi.org/10.3565/hjrp-b141}). The ARDC is enabled by the National Collaborative Research Infrastructure Strategy (NCRIS).

The authors declare that they do not have any competing interests.
\end{ack}

\bibliographystyle{plainnat}
\bibliography{references}

\appendix
\section{Pre-registration Deviations, Incidents, and Reproducibility}
\label{app:deviations}

The design for protest and assassination was pre-registered before any model labelled a text in the main annotation run. The other four concepts were pre-registered after the protest and assassination results were known and before any model annotated them.

\paragraph{Changes to the pre-registered analyses.}
The pre-registration defines a disagreement rate for each attribute on which two definitions conflict, such as the minimum size of a crowd. This rate and its version normalised by the paraphrase floor were to be computed on the texts from outside the datasets of the two definitions. Instead, this paper reports definition sensitivity for each concept on all analysed texts, as a gap and a ratio against the paraphrase floor. We also did not test the three pre-registered hypotheses on the normalised rate or run the robustness bootstrap that resamples whole articles.

\paragraph{How Sonnet was run.}
Sonnet ran through an agentic command-line harness with the pre-registered prompts and response format. For protest and assassination, each conversation contained one batch of at most 50 texts and one prompt. For the other four concepts, one conversation contained up to seven batches of one concept in turn, each with its own prompt. Because the harness lacks temperature control, the repeat floor of Section~\ref{sec:design} replaces the pre-registered re-run at temperature 0.\ifsonnetbreadth{} We added Sonnet's labels for the other four concepts after the analysis of all other labels was complete. On repeat batches of terrorism, hate speech, and climate contrarianism, Sonnet reproduced \taeRepeatBreadthOkAgr\% of \taeRepeatBreadthOkN{} annotations.\fi{}

\ifsonnet
Sonnet has a setting, the reasoning effort, that controls how much the model reasons before it answers. Sonnet labelled the texts of each prompt at one effort, except for the prompt with the original wording of the ACLED definition. For this prompt, Sonnet labelled part of the texts at the unrecorded default of the harness and the rest at an effort that we set. On average, the labels from this prompt disagree with those from its two paraphrases on \taeEffortMixedPct\% of the texts. The two paraphrases disagree with each other on \taeEffortCleanPct\% of the texts, so the two efforts do not raise Sonnet's paraphrase floor for protest.
\fi

\ifrerun
\paragraph{Qwen's re-run on the other four concepts.}
\label{app:dev-rerun}
We first ran Qwen on the other four concepts with a newer version of the inference software than the pre-registration specified. In this run, a grammar restricted Qwen's output to the response format, in place of the pre-registered JSON mode, which only requires valid JSON. With this grammar, Qwen scored \taeGTwoBreadthQwenFirst{} of \taeGTwoBreadthDen{} questions in the comprehension check. Most of the answers that it missed were truncated at the length limit that the grammar set for each answer field. We then re-ran the check and the annotation with the pre-registered version of the inference software and JSON mode, and Qwen scored \taeGTwoBreadthQwen{}. Section~\ref{sec:results} reports this re-run. Compared with the first run, the re-run changed \taeRerunDisagreePct\% of \taeRerunN{} annotations and the definition sensitivity of each concept by at most \taeRerunMaxShift{} percentage points.
\fi

\paragraph{Texts that were excluded.}
We excluded \taeNCorrupt{} assassination texts, because our download returned binary files without readable text. To find the grey-zone cases, our code extracted from each text the attributes on which the definitions disagree, such as the size of a crowd. For a subset of candidate texts, Sonnet recorded the attributes that the code could not extract. We also excluded one text when we selected these cases, because the model API refused twice to process it. Among the analysed texts, \taeMissingN{} labels across \taeMissingPrompts{} prompts are missing, because Qwen returned incomplete responses. As pre-registered, we excluded every comparison that includes one of these labels.

\paragraph{Reproducibility.}
\label{app:repro}
We froze the text sample and every prompt and then saved their hashes before the main annotation. Before it added labels to a results file, our code checked whether the recorded hashes matched those of the frozen prompt and sample. We also recorded the bootstrap seed, the versions of Gemma and Qwen, and software versions (e.g.\ vLLM). To run Gemma and Qwen for protest and assassination, one Nvidia A100 GPU with 80 GB of memory was used for each model.

Our LLM output checks rejected any value other than the \Inc{}, \Exc{}, and \Indet{} labels. The label was recorded as missing after all repeated requests failed. To find errors in the analysis code, a verification script with its own code recomputed five reported values from Gemma's raw labels after the annotation. For protest, these values are definition sensitivity, the paraphrase floor, the \Inc{} share with ACLED, and the agreement between CCC 2021 and CCC 2024. For assassination, the fifth value is the share of texts labelled \Inc{} with GTD with successful attacks only that are also labelled \Inc{} with GTD. Each recomputed value agreed with the analysis tables to nine decimal places. The verification script also checked that each results file contains as many labels as the frozen sample has texts. All \taeVerifierChecks{} checks pass on the analysis tables that we report.

\paragraph{What is released.}

Our public repository (\url{https://github.com/octner/nips-tae-definition-sensitivity}) contains the code, analysis tables, and the hash of each prompt with citations of its source codebooks. The news-derived texts are licensed and, therefore, are not released. Researchers with access to ACLED, CCC, and the news URLs indexed by GDELT can rebuild a comparable corpus with the sampling design of Section~\ref{sec:design-sample}.

\end{document}